\documentclass[10pt,twocolumn,letterpaper]{article}

\usepackage{cvpr}              

\usepackage{graphicx} \usepackage{amsmath} \usepackage{amssymb}
\usepackage{booktabs} \usepackage{siunitx} \usepackage{booktabs,multirow}
\usepackage{microtype,physics} \usepackage{float}

\usepackage[pagebackref,breaklinks,colorlinks]{hyperref}

\usepackage[capitalize]{cleveref} \crefname{section}{Sec.}{Secs.}
\Crefname{section}{Section}{Sections} \Crefname{table}{Table}{Tables}
\crefname{table}{Tab.}{Tabs.}

\begin{document}
	
	\title{Real-time Hyperspectral Imaging in Hardware via Trained Metasurface Encoders}

	\author{Maksim Makarenko$^{1\dag}$, Arturo Burguete-Lopez$^{1}$, Qizhou Wang$^{1}$, Fedor Getman$^{1}$,
		Silvio Giancola$^{1}$, \\Bernard Ghanem$^{1}$, Andrea Fratalocchi$^{1}$\\ $^{1}$King
		Abdullah University of Science and Technology (KAUST)\\ Thuwal, 23955-6900,
		KSA\\ {\tt\small $^{\dag}$maksim.makarenko@kaust.edu.sa} }
	
\maketitle

\begin{abstract} Hyperspectral imaging has attracted significant attention to
	identify spectral signatures for image classification and automated pattern
	recognition in computer vision. State-of-the-art implementations of snapshot
	hyperspectral imaging rely on bulky, non-integrated, and expensive optical
	elements, including lenses, spectrometers, and filters. These macroscopic
	components do not allow fast data processing for, \eg real-time and
	high-resolution videos. This work introduces Hyplex\texttrademark{}, a new
	integrated architecture addressing the limitations discussed above.
	Hyplex\texttrademark{}  is a CMOS-compatible, fast hyperspectral camera that
	replaces bulk optics with nanoscale metasurfaces inversely designed through
	artificial intelligence. Hyplex\texttrademark{} ~does not require spectrometers
	but makes use of conventional monochrome cameras, opening up the possibility for real-time and
	high-resolution hyperspectral imaging at inexpensive costs.
	Hyplex\texttrademark{} exploits a model-driven optimization, which connects the
	physical metasurfaces layer with modern visual computing approaches based on
	end-to-end training. We design and implement a prototype version of
	Hyplex\texttrademark{} and compare its performance against the state-of-the-art
	for typical imaging tasks such as spectral reconstruction and semantic
	segmentation. In all benchmarks, Hyplex\texttrademark{} reports the smallest
	reconstruction error. We additionally present what is, to the best of our
	knowledge, the largest publicly available labeled hyperspectral dataset for
	semantic segmentation.
	\footnote{Dataset available on \href{https://github.com/makamoa/hyplex}{https://github.com/makamoa/hyplex}.}
\end{abstract}


\begin{figure}[ht] \includegraphics[width=\linewidth]{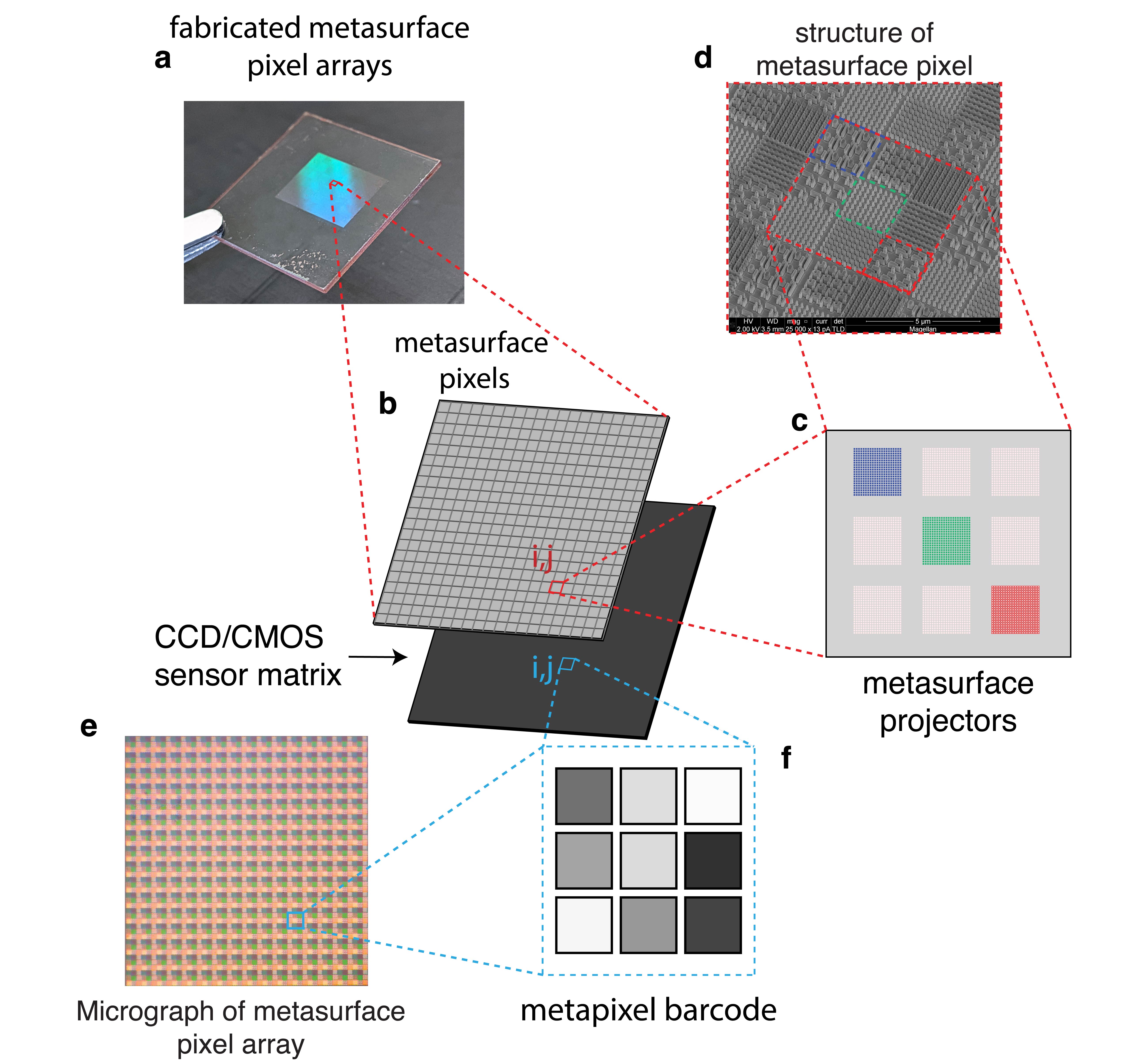}
	\caption{\textbf{Hardware implemented Hyplex\texttrademark{}~ imaging system.}
		(a) Example of metasurface pixel arrays (blue squares). (b) Schematic of
		meta-pixel array on top of a camera sensor. (c) Closeup showing the metasurface
		projectors as subpixels of the array. (d) Scanning electron microscope image of
		a fabricated metasurface pixel. (e) Optical micrograph of the metasurface pixel
		array. (f) Illustration of the barcode generated by (e). } \label{fig:concept}
\end{figure}

\section{Introduction} Hyperspectral imaging is gaining considerable interest in
many areas including civil, environmental, aerial, military, and biological
sciences for estimating spectral features that allow the identification and
remote sensing of complex materials~\cite{HyperAll, HyperAgriBook}. Ground-based
hyperspectral imaging enables automated classification for food inspection,
surgery, biology, dental and medical
diagnosis~\cite{Afromowitz1988,Panasyuk2007,Lu2014,Gowen2015}. Likewise, aerial
and submarine hyperspectral imaging are currently opening new frontiers in
agriculture and marine biology for the taxonomic classification of fauna, and
through aerial drone footage for precision agriculture~\cite{HyperAgri,HyperAgriBook,Chennu2017,HyperWater}. The present state-of-the-art in
hyperspectral imaging, however, is still affected by problems of expensive setup
costs, time-consuming post-data processing, low speed of data acquisition, and
the needs of macroscopic optical and mechanical
components~\cite{Oiknine2019,Yoon2011}. A single hyperspectral image obtained
from a high-resolution camera typically requires gigabytes of storage space,
making it impossible to perform real-time video analysis with today's computer
vision techniques~\cite{Kehtarnavaz2006a}.

Computational hyperspectral reconstruction from a single RGB image is a promising
technique to overcome some of the challenges mentioned
above~\cite{rgb1,arab1,Jia_2017_ICCV,galliani2017learned,Alvarez_Gila_2017,Xiong2017HSCNNCH,Jiang2013WhatIT,nguyen2014training,He2018,zhao2020hierarchical}.
 Heidrich \etal \cite{HyperspectralDOE:SIG:2019} proposed hyperspectral cameras based on integrated 
diffractive optical elements, while other groups~\cite{Yu2014,Chen2018} leveraged deep neural networks 
for designing spectral reconstruction filters. While these approaches could help address the problem of 
speed, they are not yet able to tackle the issues of high cost and slow data processing. Other 
bottlenecks are the use of elementary filter responses, which are not optimized beyond primitive 
thin-film interference patterns, and the lack of integrated structures that could exploit the modern 
footprint of CCD/CMOS sensors.

We here introduce the Hyplex\texttrademark{} system (\cref{fig:concept}), a
data-driven hyperspectral imaging camera (\cref{fig:concept}, a-b), which uses
state-of-the-art metasurfaces to replace macroscopic components with highly
integrated dielectric nanoresonators that manipulate light as a feed-forward
neural network~\cite{Getman2021,Galinski2017a,Bonifazi2020}. Metasurfaces have
successfully demonstrated the ability to integrate various basic optical
components for different
applications~\cite{tittl2018imaging,Tittl1105,Tittl2019}. Hyplex\texttrademark{}
leverages this technology to compress high-dimensional spectral data into a
low-dimensional space via suitably defined projectors (\cref{fig:concept}, c-d),
designed with end-to-end learning of large hyperspectral datasets.
ALFRED~\cite{Getman2021,Fratalocchia,Makarenko2021robust}, an open-source,
inverse-design software exploiting artificial intelligence (AI), provides the
means to design the metasurface projectors. These nanostructures encode
broadband information carried by incoming spectra into a barcode composed of a
discrete pattern of intensity signals  (\cref{fig:concept}, e-f). A physical
model-aware framework finds the optimal projectors' response with various
learning schemes, designed based on user end tasks. %
	
We summarize our contribution as follows: 
    \textbf{(i)} We propose and
	implement an inexpensive and fast-processing data-driven snapshot hyperspectral
	camera that uses two integrated components: inverse-designed spectral encoders
	and a monochrome camera. \textbf{(ii)} We implement an end-to-end framework for
	hyperspectral semantic image segmentation and spectral reconstruction, and
	benchmark it against the state-of-the-art, reporting the highest
	performance to date. \textbf{(iii)} We create FVgNET, the largest publicly available
	dataset of $317$ samples of labeled hyperspectral images for semantic segmentation
	and classification.

\section{Related Work} Hyperspectral reconstruction is an ill-posed problem
demanding the inverse projection from low-dimensional RGB images to densely
sampled hyperspectral images (HSI)~\cite{arad2016sparse,nie2018deeply}.
Metamerism~\cite{foster2006frequency}, which projects different spectral
distributions to similar activation levels of visual sensors, represents a
significant challenge. Traditional RGB cameras project the entire visible
spectra into only three primary colors. This process eliminates critical
information making it challenging to distinguish different objects
\cite{nguyen2014training}. For the specific task of hyperspectral
reconstruction, we can partially recover such lost information. Spectral
projections are similar to autoencoders in the sense that they downsample the
input to a low-dimensional space. If we design a suitable algorithm that
explores this space efficiently, we could retrieve sufficient data to
reconstruct the initial input.

\noindent\textbf{Reconstruction by sparse coding and deep learning:} Sparse
coding \cite{lee2007efficient,robles2015single} represents perhaps the most
intuitive approach to this idea. These methods statically discover a set of
basis vectors from HSI datasets known \emph{a priori}. Arad
\etal~\cite{arad2016sparse} implemented the K-SVD algorithm to create
overcomplete HSI and RGB dictionaries. The HSI is reconstructed by decomposing
the input image into a linear combination of basis vectors, then transferred
into the hyperspectral dictionary. A limit of sparse-coding methods is their
applied matrix decomposition algorithms, which are vulnerable to outliers and
show degraded performance~\cite{kawakami2011high}.
Recently, research groups extended the capabilities of sparse coding by
investigating deep learning. Galliani \etal~\cite{galliani2017learned}
demonstrated a supervised learning method, where a UNet-like architecture
\cite{ronneberger2015u} is trained to predict HSI out of single RGB images.
Nguyen~\cite{nguyen2014training} trained a radial basis function network to
translate white-balanced RGB values to reflection spectra. In another work,
Xiong \etal~\cite{xiong2017hscnn} introduced a 2-stage reconstruction approach
comprising an interpolation-based upsampling method on RGB images. The
end-to-end training proposed recovers true HSI from the upsampled images. Wug
\etal~\cite{oh2016yourself} used different RGB cameras to acquire
non-overlapping spectral information to reconstruct the HSI. These approaches
reconstruct spectral information from highly non-linear prediction models,
limited by their supervised learning structure. The models constrain data
downsampling to non-optimal RGB images by applying a color generation function
on HSI or generic RGB cameras. With Hyplex\texttrademark{}, we avoid all the
issues of the sparse coding and deep-learning reconstruction methods by
exploring a new concept, which performs spectral downsampling with optimally
designed metasurface projectors.

\noindent\textbf{Hyperspectral imaging with trainable projectors:} Optical
projectors in cameras mimic the chromatic vision of humans based on primary
colors~\cite{ibraheem2012understanding}. In hyperspectral imaging, however, the
design of projectors requires further study to identify their optimal number and
response. Human eyes are not the best imaging apparatus for every possible
real-world scenario. The works
of~\cite{Arad2017FilterSF,shen2014channel,wu2019optimized} expand the concept of
RGB cameras to arbitrary low-dimensional sampling of reflectance spectra. These
works employ different variants of optimization routines, which converge to a
set of optimal projectors from an initial number of candidates. The selected
projectors provide a three-channel reconstruction of the HSI with superior
performance. Nie \etal~\cite{nie2018deeply} demonstrated that a 1$\times$1
convolution operation achieves similar functionality to optical projectors while
processing multi-spectral data frames. The network is like an autoencoder, where
the input HSI is downsampled and then reconstructed by a decoder network. Zhang
\etal~\cite{zhang2021deeply} designed and fabricated a broadband encoding
stochastic camera containing 16 trainable projectors that map high-dimensional
spectra to lower-dimensional intensity matrices. Recently, Liutao
\etal~\cite{yang2021fs} proposed FS-Net, a filter-selection network for
task-specific hyperspectral image analysis. In~\cite{yu2021optical}, the authors
showcased an idea of filter optimization for hyperspectral-informed image
segmentation tasks.

\noindent\textbf{Inverse design of metasurface projectors:} Optimizing best-fit
filters is a dimensionality reduction problem, which requires finding the
principal component directions that encode eigenvectors showing the lowest loss.
The state-of-the-art results are generated either from theoretical calculation
or experimental measurement on thin-film filters, representing a rough
approximation of the precise principal components. In hyperspectral imaging,
these components typically exhibit frequency-dependent irregular patterns
composed of complex distributions of sharp and broad resonances, indicating the
need for more dedicated control of material structures, \eg metasurface
technology. Modern metasurface design approaches \cite{tseng2021neural, wang2021advancing} usually rely on a library of pre-computed metasurface responses and polynomial fitting to  further generalize the relationship between design parameters and the device performance. We, instead, design our metasurface optical projectors via ALFRED \cite{getman2021broadband}, a hybrid inverse design scheme that combines
classical optimization and deep learning\cite{wang2021advancing}. 
In this work, we significantly extend the capabilities of the original code by adding differentiability, physical-model regularization, and complex decoder projectors able to tackle different computer vision tasks and perform thousands of parameter optimizations through the supervised end-to-end learning process.


\section{Methodology} \label{sec:method} 

\begin{figure}[ht]
	\includegraphics[width=\linewidth]{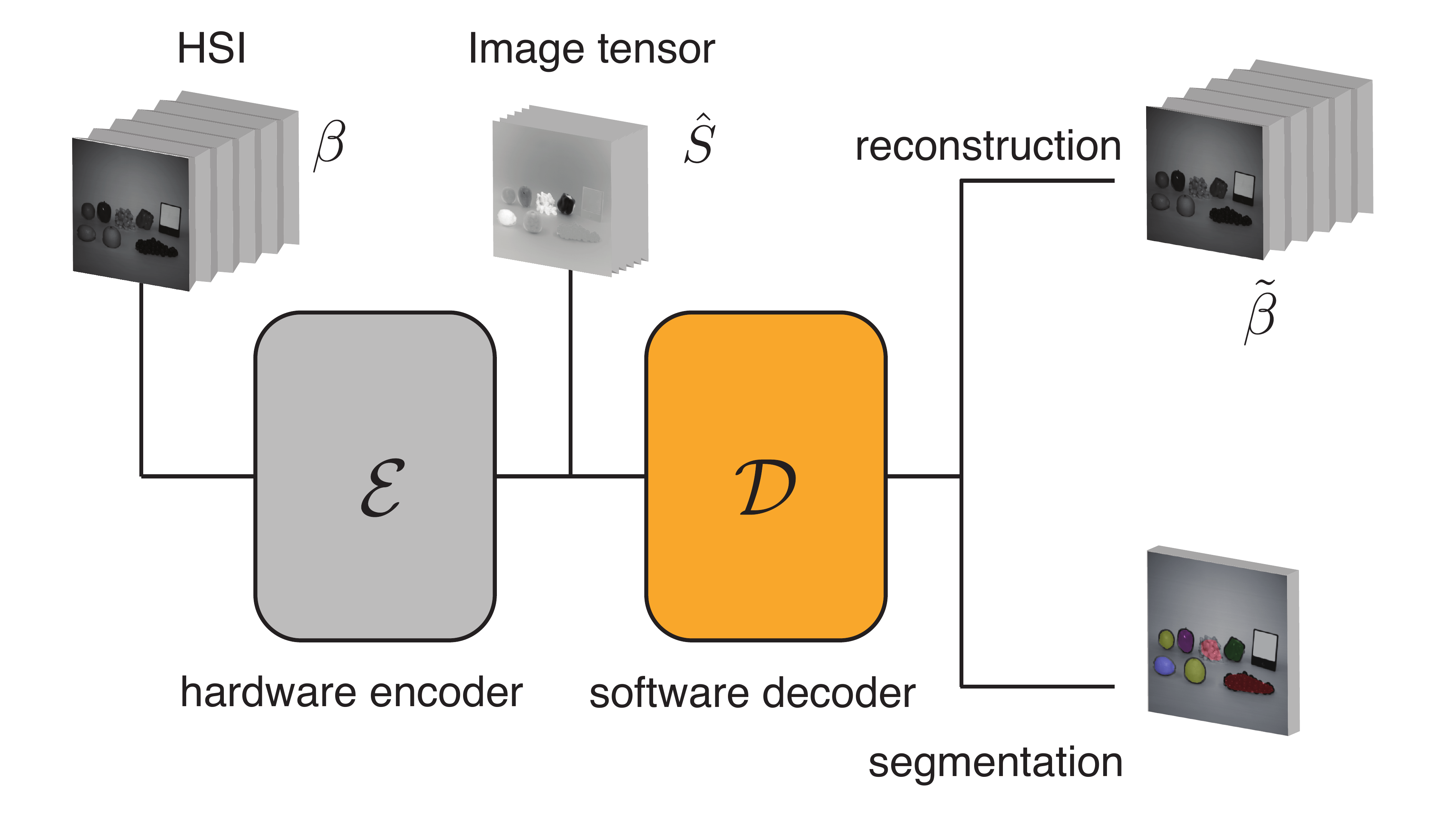}
	\caption{\textbf{Conceptual sketch of Hyplex\texttrademark{} system.} The system
		is constructed by a hardware optical encoder $\mathcal{E}$ that is implemented
		via trainable metasurface arrays and a software decoder $\mathcal{D}$ optimized
		for two different tasks, including hyperspectral reconstruction and
		spectral-informed semantic segmentation.} \label{fig:encoder} \end{figure}

The Hyplex\texttrademark{} hyperspectral imaging system consists of two parts: a
hardware linear spectral encoder $\mathcal{E}$ and a software decoder
(\cref{fig:encoder}). The encoder compresses an input high-dimensional HSI
$\boldsymbol{\beta}$ to a lower multispectral image tensor
$\hat{S}=\mathcal{E}(\beta)$, while the decoder maps the tensor $\hat{S}$ to
user-defined task-specific outputs. In this work, we consider two types of
tasks: hyperspectral reconstruction and semantic segmentation. Spectral
reconstruction aims to reconstruct with minimum losses the input HSI tensor. We
define the loss via the Root Mean Squared Error (RMSE)
$\hat{\beta}=\mathcal{D}_{rec}(\mathcal{E}(\beta))$ between reconstructed and
input spectra. Semantic segmentation, conversely, provides a pixel-by-pixel
classification of HSI. In this task, we use as decoder $\mathcal{D}_{seg}$ the
U-Net architecture, with adjusted input and output layers to meet the
dimensionality of the HSI tensor. The decoder outputs softmax logits $\hat{y}$,
representing the probability of observing each pixel ground-truth label $y$. We
assess these predictions quantitatively by using the Cross-Entropy loss function
$\mathcal{L}_{seg}$.

\subsection{Hardware encoder} \label{sec:encoder} Recent work demonstrates that
the transfer function of an array of sub-micron nanostructured geometries can
approximate arbitrarily defined continuous functions
\cite{Makarenko2020,Getman2021}. In Hyplex\texttrademark{}, we exploit such
universal approximation ability to design and implement an optimal linear
spectral encoder hardware for a specific hyperspectral information-related
imaging task. 

\begin{figure*}[ht] \centering
	\includegraphics[width=\textwidth]{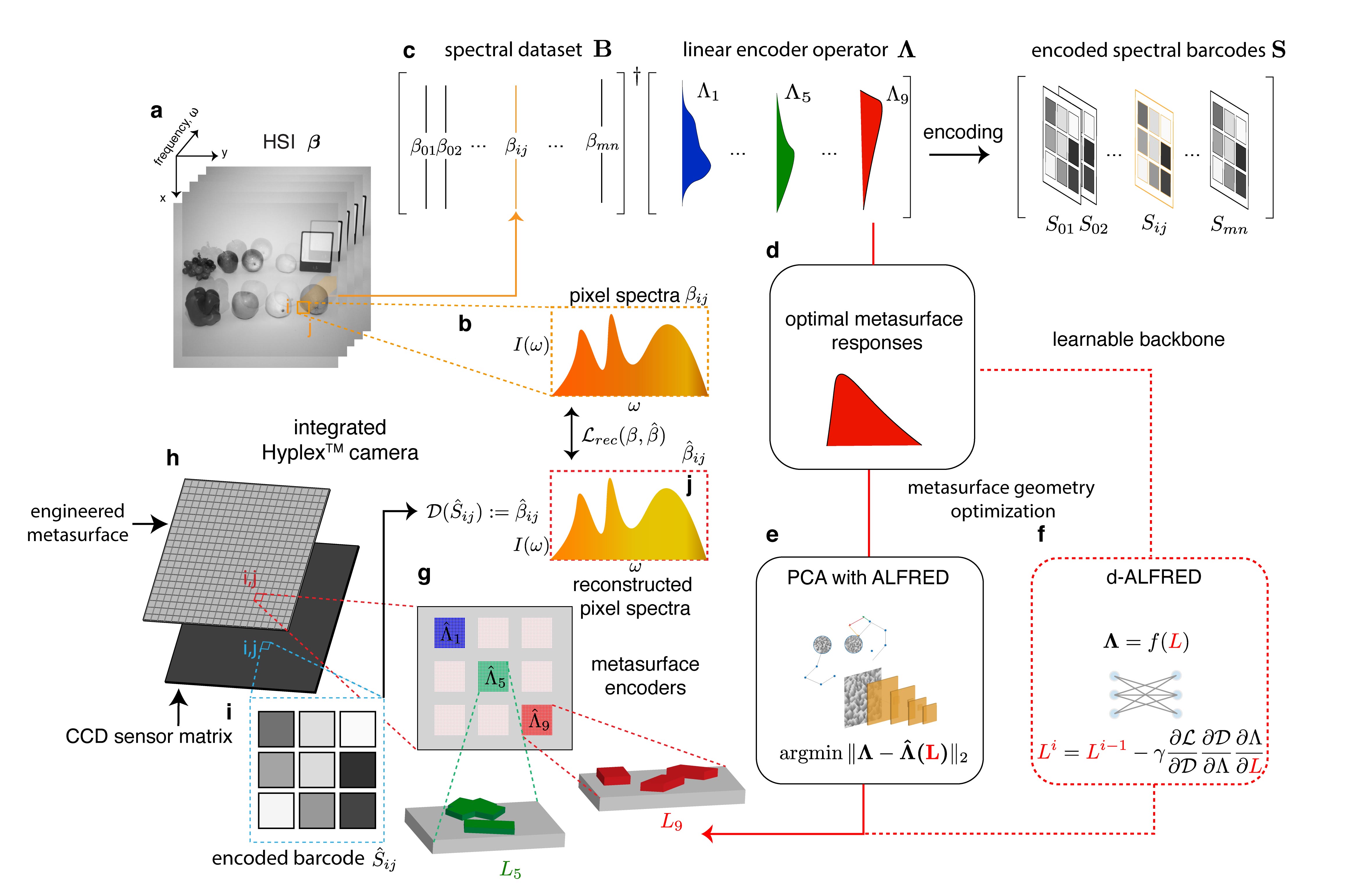} \caption{
		\textbf{Metasurface subpixel array as a linear spectral encoder.} (a) A spectral
		image tensor ($\boldsymbol{\beta}$) is captured by a hyperspectral camera. (b)
		The corresponding pixel spectra ($\hat{\beta_{ij}}$) at position $ij$ in the
		$xy$ camera plane. (c) Example of dimensional reduction linear operator
		$\boldsymbol{\Lambda}^\dag$ of a flattened matrix $\mathbf{B}$ with the resulted
		projected encoder barcode for a pixel spectra at the $ij$ position. (d) Optimal
		encoder functions $\boldsymbol{\Lambda}^\dag$. (e) Non-differentiable inverse
		design optimization framework implemented via ALFRED utilized to find a set of
		metasurfaces $\mathbf{L}$ with desired response $\Lambda_i$. (f) Differentiable
		backbone enabling simultaneous optimization of responses $\boldsymbol{\Lambda}$
		and structures $\mathbf{L}$. Metasurface pixel (g) composed of a two-dimensional
		array of resonant metapixels with corresponding fitted transmission responses
		$\hat{\boldsymbol{\Lambda}}$. (h) Conceptual sketch of the
		Hyplex\texttrademark{} system with an enlarged spectral-specific barcode (i)
		produced by an imaging-based readout of the metasurface’s transmission
		response.(j) Recovered pixel spectra through decoder $\mathcal{D}_{rec}$
		projection $\hat{\beta}_{ij}$.} \label{fig:pca} \end{figure*}

\Cref{fig:pca} summarizes the data workflow of Hyplex\texttrademark{} for a
generic linear encoder operator $\mathcal{E}=\boldsymbol{\hat{\Lambda}}^{\dag}$.
Panel (a) shows an example hyperspectral image. The data is represented as a
tensor $\boldsymbol{\beta}$ with three dimensions: two spatial dimensions
$(x,y)$, corresponding to the camera virtual image plane, and one frequency axis
$\omega$, measuring the power density spectra retrieved at one camera pixel
(\cref{fig:pca}b). Following a data-driven approach, we implement a linear
dimensionality reduction operator that finds a new equivalent encoded
representation of $\boldsymbol{\beta}$ (\cref{fig:pca}c). The hyperspectral
tensor of a dataset of images is flattened to a matrix $\mathbf{B}$ that
contains, on each column, the power density spectra of a set of camera pixels.
We then the apply the linear encoding $\boldsymbol{\Lambda}^{\dag}$ to obtain an
approximation of $\mathbf{B}$ \cite{bishop} via a set of linear projectors
$\boldsymbol{\Lambda}(\omega)$, which map pixel-by-pixel the spectral coordinate
$\beta_{ij}$ to a set of scalar coefficients $S_{ijk}$: 

\begin{equation}
	\label{eq:projection} S_{ij} =
	\boldsymbol{\tilde{\Lambda}}(\omega)\beta{ij}(\omega),\;\;\; S_{ijk} = \int
	\Lambda_k(\omega) \beta_{ij}(\omega) \dd{\omega}. \end{equation} 
	
The spectral
information contained in $\beta_{ij}(\omega)$ is embedded into an equivalent
barcode $S_{ijk}$ of a few components. To implement the $\boldsymbol{\Lambda}$
encoder projectors into hardware, Hyplex\texttrademark{} uses two different
engineering lines (\cref{fig:pca}e-f). When the user end task does not require
additional constraints, such as in, \eg spectral reconstruction,
Hyplex\texttrademark{} implements the projector by utilizing optimization
frameworks to minimize the norm between the physical metasurface response
$\hat{\boldsymbol\Lambda}$ and the target $\boldsymbol{\Lambda}$
(\cref{fig:pca}e). Conversely, in tasks that require further conditions such as,
\eg hyperspectral semantic segmentation, Hyplex\texttrademark{} uses a
learnable backbone (\cref{fig:pca}f). This optimization exploits d-ALFRED, a new
version of ALFRED that creates a differentiable physical model that is trained
with an end-to-end approach. d-ALFRED designs metasurface geometries with an
iterative process that minimizes the loss function $\mathcal{L}_{seg}$ by
optimizing simultaneously the projector responses $\boldsymbol{\Lambda}$ and the
vector $\mathbf{L}$ containing all the parameters defining the metasurface. A
single Hyplex\texttrademark{} pixel (\cref{fig:pca}g) integrates various
metasurface projectors in a two-dimensional array of sub-pixels, which are
replicated in space to form the Hyplex\texttrademark{} hardware encoder
(\cref{fig:pca}h). The encoder transforms a reflection spectra arising from a
scene into a barcode $\hat{S}_{ij}$ (\cref{fig:pca}i), composed of a set of
intensity signals proportional to the overlap between the input spectra and each
projector's response as defined in \cref{eq:projection}. A standard
monochromatic camera, placed behind the metasurfaces, acts as an imaging readout
layer. Each pixel of the camera matches the sub-pixel of the hardware encoder
and retrieves one intensity signal of the barcode $\hat{S}_{ij}$
(\cref{fig:pca}j).

\noindent\textbf{PCA projectors engineered with ALFRED:} We use a linear encoder
$\boldsymbol{\Lambda}$ obtained through an unsupervised learning technique via
principal component analysis (PCA). The PCA performs hardware encoding
$\mathcal{E}$ by selecting the $k$ strongest ($k=9$ for this work) principal
components $\boldsymbol{\tilde{\Lambda}^\dag}$ from the singular value
decomposition of
$\mathbf{B}=\boldsymbol\Lambda\boldsymbol\Sigma\mathbf{V}^\dag$~\cite{bishop},
and approximating $\mathbf{B}$ as follows: 

\begin{equation} \label{uno}
	\mathbf{B}\approx
	\boldsymbol{\tilde{\Lambda}}\boldsymbol{\tilde{\Sigma}}\pmb{\tilde{V}}^\dag
\end{equation} 

\Cref{uno} offers the closest linear approximation of
$\mathbf{B}$ in least square sense. We implement the decoder $\mathcal{D}$ with
the linear projector
$\hat{\beta}_{ij}=\boldsymbol{\tilde{\Lambda}}\hat{S}_{ij}$, which recovers the
best least square approximation of the pixel spectra $\hat{\beta}_{ij}(\omega)
\approx \beta_{ij}(\omega)$ (\cref{fig:pca}j) from the selected PCA component.

\subsection{Learnable backbone via differentiable physical model}
\label{sec:backbone} 
In this approach, we represent the decoder operator $\mathcal{D}$ as a set of
hierarchical nonlinear operators $\mathcal{F}$, which project the input tensor
$\hat{S}$ into an output measurement tensor $\hat{y}$. This process is
iteratively trained via supervised learning, comparing the measurement $\hat{y}$
with some ground-truth tensor $\tilde{y}$. This end-to-end training finds the
optimal feature space $\hat{S}$ and the associated linear projectors
$\boldsymbol{\Lambda}$. To train Hyplex\texttrademark{} in this framework with
backpropagation, the encoder $\mathcal{E}$ needs to be differentiable.

In the inverse-design of projectors, the encoder $\mathcal{E}=\mathbf{H}$, with
$\mathbf{H}(\omega)$ representing the output transmission function of the
metasurface response, which is obtained from the solution of the following set
of coupled-mode equations~\cite{Makarenko2020}: \begin{equation} \label{cme}
	\left\{\begin{array}{l}
		\tilde{\mathbf{a}}(\omega)=\frac{\tilde{K}}{i(\omega-W)+\frac{\tilde{K}
				\tilde{K}^{\dagger}}{2}} \tilde{\mathbf{s}}_{+} \\
		\tilde{\mathbf{s}}_{-}(\omega)=\tilde{C}(\omega)
		\cdot\left(\tilde{\mathbf{s}}_{+}-\tilde{K}^{\dagger} \cdot
		\tilde{\mathbf{a}}\right) \end{array}\right. \end{equation} where $W$ is a
diagonal matrix with resonant frequencies $\omega_{n}$ of the modes $W_{n n}=$
$\omega_{n}$, $\tilde{C}(\omega)$ is a scattering matrix modeling the scattering
of impinging waves $\tilde{\mathbf{s}}_{+}$ on the resonator space, and
$\tilde{K}$ is a coupling matrix representing the interaction between traveling
waves $\tilde{\mathbf{s}}_{\pm}(t)$ and resonator modes $\tilde{\mathbf{a}}(t)$.
Equations (\ref{cme}) describe the dynamics of a network of  resonator modes
$\tilde{\mathbf{a}}=\left[\tilde{a}_{1}(\omega), \ldots,
\tilde{a}_{n}(\omega)\right]$, interacting with
$\tilde{\mathbf{s}_{\pm}}=\left[\tilde{s}_{1 \pm}(\omega), \ldots, \tilde{s}_{m
	\pm}(\omega)\right]$ incoming $(+)$ and reflected $(-)$ waves. Section 1 of the
Supplementary Material provides more details on the quantities appearing in
\cref{cme}.

\begin{figure}[htbp] \includegraphics[width=\linewidth]{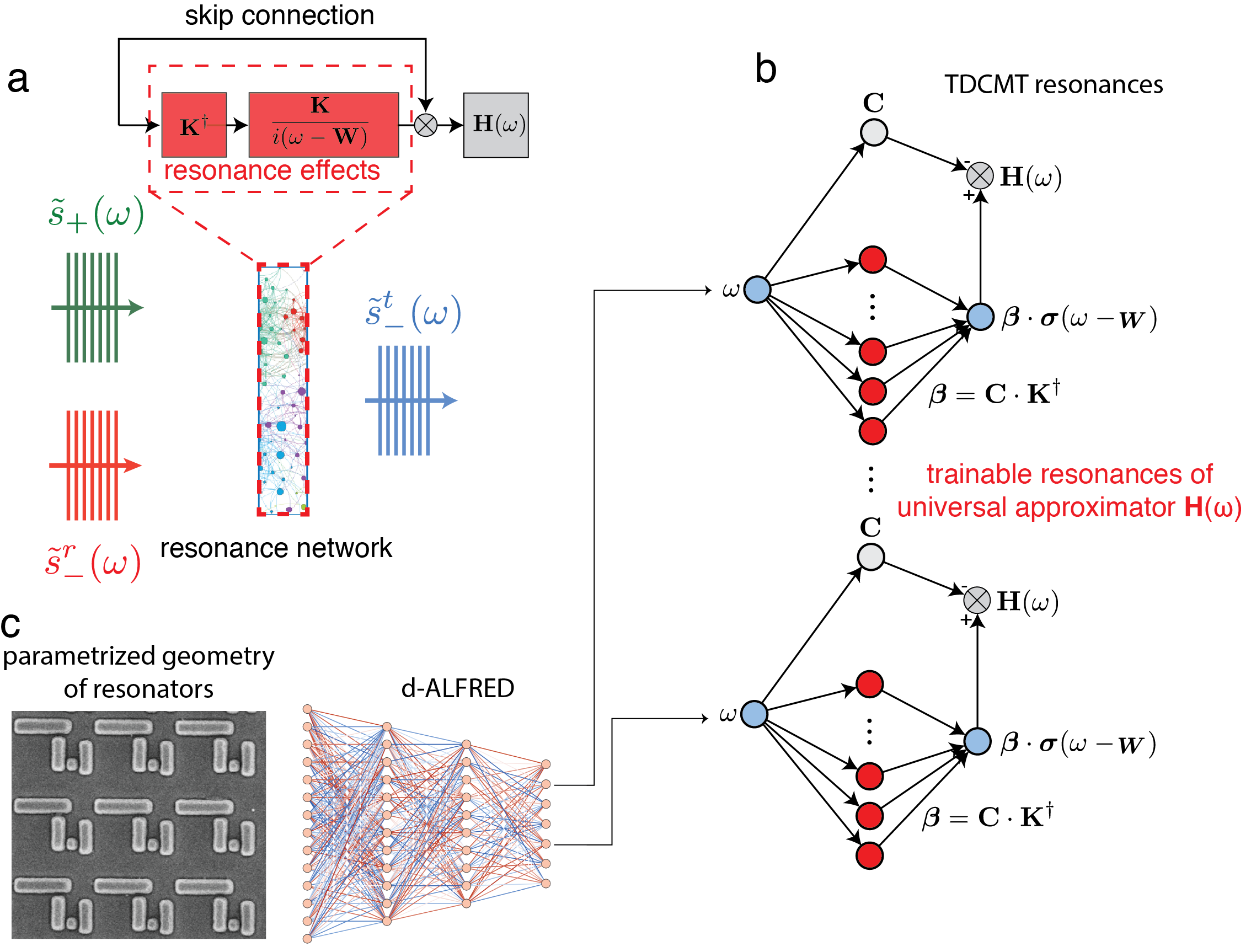}
	\caption{\textbf{Coupled mode network as a differentiable metasurface physical
			model.} (a) Coupled-mode photonic network as a feedback-loop with skip
		connection. (b) trainable coupled resonance layer. (c) d-ALFRED: trained
		differentiable projections from parametric geometry shapes to resonances.}
	\label{fig:tdcmt} \end{figure} 
	
	The input-output transfer function
$\mathbf{H}=\tilde{\mathbf{s}}_{-}/\tilde{\mathbf{s}}_{+}$ resulting from the
solution of \cref{cme} is the superposition of two main terms: a propagation
term defined by the scattering matrix $\tilde{C}(\omega)$ and a nonlinear term
containing the rational function $\frac{\tilde{K}}{\sigma(\omega - W)}$.
\Cref{cme} represents a differentiable function of $W$ through which it is
possible to backpropagate (\cref{fig:tdcmt}~b).

\noindent\textbf{d-ALFRED:} To project the resonator quantities in \cref{cme} to
metasurface input parameters $\mathbf{L}$, we use a supervised optimization
process. We train a deep neural network to learn the relationship between
$\mathbf{L}$ and the resonator variables in \cref{cme}. Following the same
approach of~\cite{Makarenko2021robust}, we train the network with a supervised
spectral prediction task by using arrays of silicon boxes with simulated transmission/reflection responses (see Sec. 2 of Supplementary Material).

\begin{figure*}[htbp] \centering
	\includegraphics[width=\textwidth]{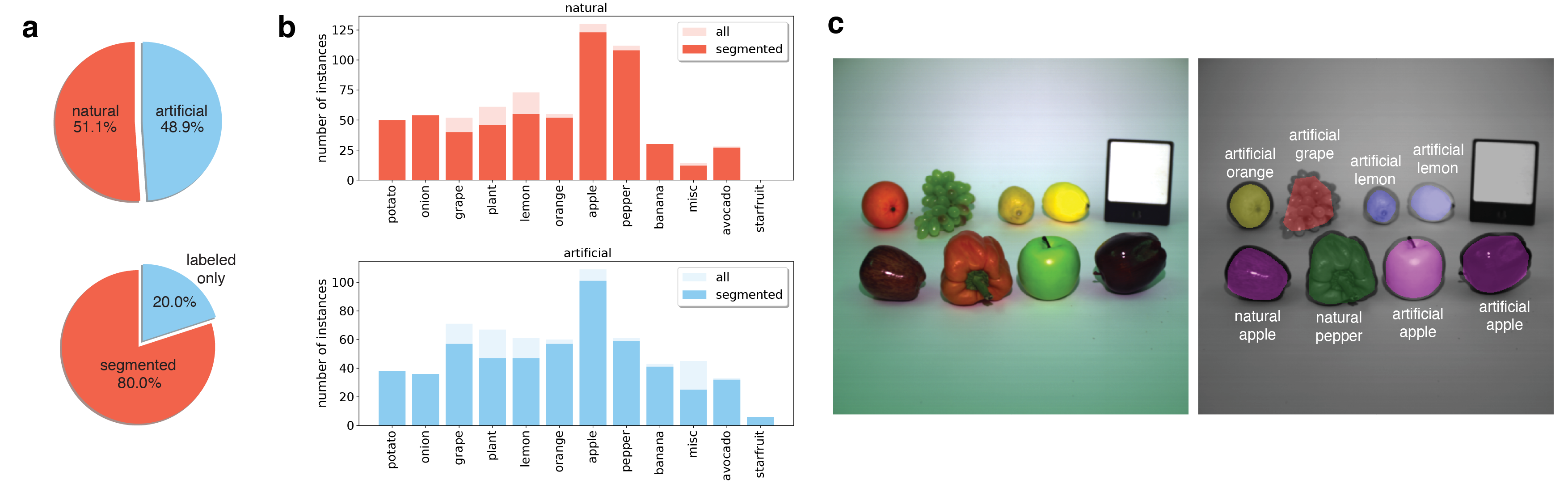} \caption{
		\textbf{Example and statistical analysis on our dataset} (a)  Overview of the
		composition of the dataset. There exist a near equal number of natural and
		artificial objects in the scenes, 80\% of the images are with segmentation masks
		and the rest with labels only. (b) Distribution of scene objects in classes.
		Each class has a roughly equal number of instances in the dataset with the
		exception of apples and peppers, as they have more chromatic variety. (c) Left:
		RGB visualization of hyperspectral image. Right: Segmentation mask and labels
		for each object. } \label{fig:dataset} \end{figure*}

\section{Datasets}\label{sec:datasets} To train and validate the
Hyplex\texttrademark{} system, we use three publicly available datasets: the
CAVE dataset, consisting of 32 indoor images covering \SIrange{400}{700}{nm},
and the Harvard and KAUST sets, which contain both indoor and outdoor scenes,
and amount to $75$ and $409$ images, respectively, with spectral bands covering
\SIrange{420}{720}{nm} and \SIrange{400}{700}{nm} respectively. We create an additional hyperspectral dataset FVgNET. FVgNET is comprised of $317$ scenes showing fruits
and vegetables, both natural and artificial, taken indoors under controlled
lighting conditions, and covering the \SIrange{400}{1000}{nm} range. We acquired
the images using a setup consisting of a white paper sheet arranged in an
infinity curve, a configuration employed in photography to isolate objects from
the background. We achieve good spectral coverage while minimizing the presence
of shadows in the final images by illuminating the objects with overhead white
LED indoor lighting, a \SI{150}{W} halogen lamp (OSL2 from Thorlabs) equipped
with a glass diffuser and a \SI{100}{W} tungsten bulb mounted in a diffuse
reflector.

\Cref{fig:dataset}a-b shows the distribution of object classes in the dataset.
For each class of objects (\eg, apple, orange, pepper), we generated an
approximately equal number of scenes showing: natural objects only and
artificial objects only. The dataset consists of $12$ classes, represented in the images proportionally to their chromatic variety.
Furthermore, we annotated $80\%$ of our images with addititional segmentation masks.
%
%
%
We incorporate semantic segmentation masks into the dataset by processing the
RGB images generated from the $204$ spectral channels. We acquired the images in such a way to avoid
the intersection of objects, allowing for automatic generation of masks for
the areas occupied by each object. We then annotated each marked object,
identifying each object class and whether they are natural or artificial.
\Cref{fig:dataset}c illustrates the implementation of the semantic segmentation mask on an image of the dataset. For more details about the FVgNET dataset please refer to Sec. 3 of Supplementary Material.

\section{Results}

\setlength\tabcolsep{8pt} \begin{table*}[!htb] \centering
	\begin{tabular}{@{}llllll@{}} \toprule
		\multicolumn{1}{c}{\multirow{2}{*}{Model}} & \multicolumn{5}{c}{Dataset} \\
		\cmidrule(l){2-6} \multicolumn{1}{c}{} &CAVE \cite{CAVE_0293} & Harvard(out)
		\cite{chakrabarti2011statistics} & Harvard(in) \cite{chakrabarti2011statistics}
		& KAUST \cite{Yuqi2021SpecSeperation} & FVgNET \\ \midrule Nguyen \etal
		\cite{nguyen2014training} &14.91$\pm$11.09 & 9.06$\pm$9.69 & 15.61$\pm$8.76 & -
		& - \\ Arad and Ben-Shahar \cite{arad2016sparse} &8.84$\pm$7.23 &
		14.89$\pm$13.23 & 9.74$\pm$7.45 & - & - \\ Jia \etal \cite{Jia_2017_ICCV}
		&7.92$\pm$3.33 &  8.72 $\pm$7.40 & 9.50$\pm$6.32 & - &- \\ Nie \etal
		\cite{nie2018deeply}  & 4.48 $\pm$ 2.97& 7.57$\pm$4.59 & 8.88 $\pm$ 4.25 & - & -
		\\ 
		Hyplex\texttrademark{} &\textbf{2.05}$\pm$ \textbf{1.82}& \textbf{2.13} $\pm$
		\textbf{1.81}  & \textbf{6.65} $\pm$ \textbf{5.88} & 2.23 $\pm$ 3.35 & 1.73
		$\pm$ 1.35 \\\bottomrule 
	\end{tabular} \caption{\textbf{Comparison of baselines.} We report the RMSE from
		spectral reconstruction in multiple hyperspectral datasets}
	\label{tab:tablemerit} \end{table*}

\subsection{Hardware implementation}

\begin{figure}[htbp] \centering
	\includegraphics[width=\linewidth]{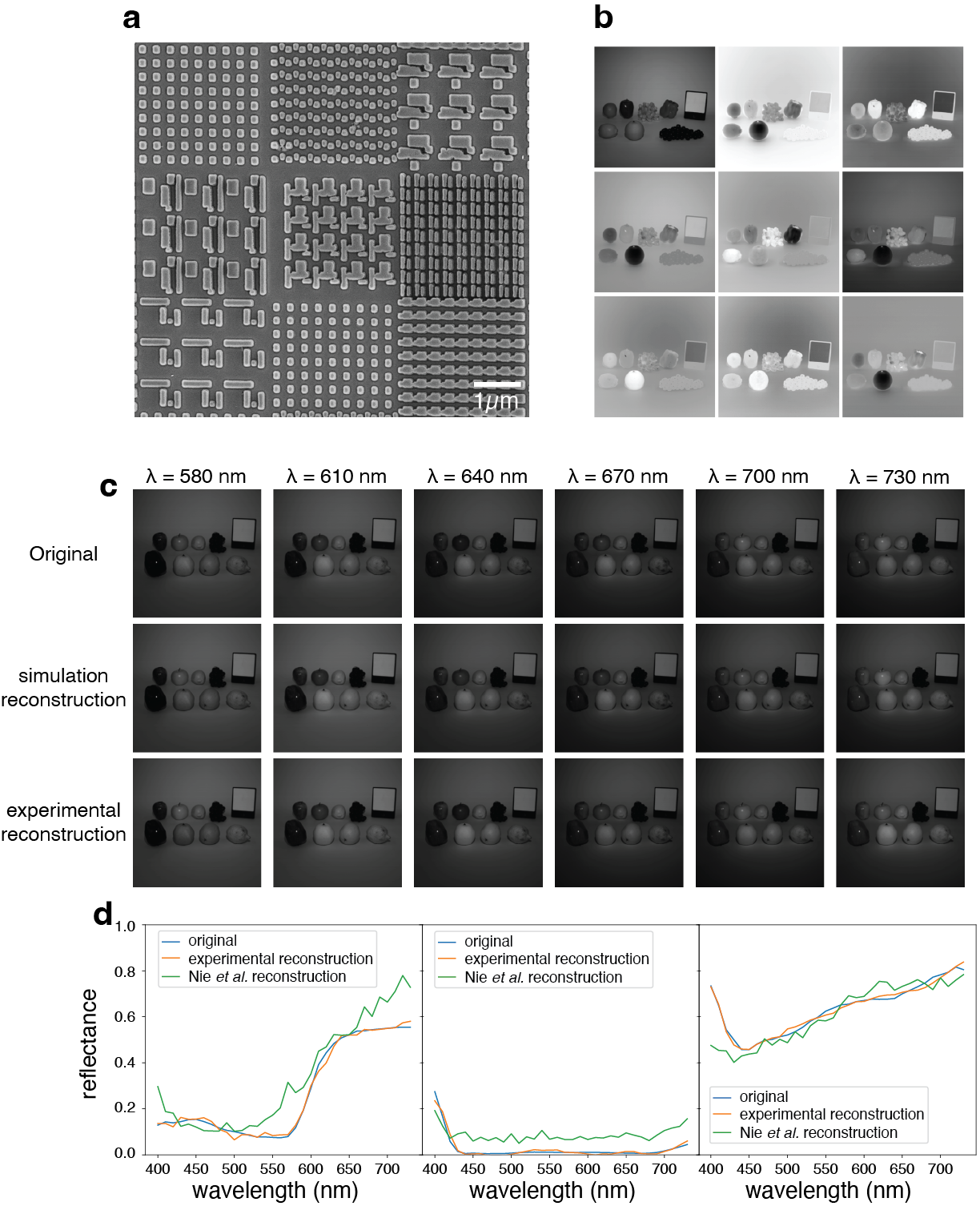}
	\caption{\textbf{Spectral reconstruction.} (a) Scanning electron microscope
		image of the array of projectors. (b) The output of the scene processed by our
		projectors. (c) Comparison between acquired and recovered hyperspectral image
		using the theoretical (middle row) and experimental (lower row) responses of our
		projectors. (d) Comparison between the original spectra and their reconstruction
		using the projectors and the reconstruction algorithm by Nie \etal.
		\cite{nie2018deeply} for random pixels of the scene in (c).}
	\label{fig:hardware} \end{figure}
	
We fabricate arrays of metasurface
projectors by patterning thin layers of amorphous silicon deposited on optical
grade fused silica glass slides. \Cref{fig:hardware}a shows a scanning electron
microscope (SEM) image of a manufactured metasurface pixel, detailing the
nanoscale structure of each of the nine projectors. We produce each projector of
the $3\times 3$ sub-array, so it occupies the area of a \SI{2.4}{\micro\meter}
wide square, a size typical for the pixels present in modern digital camera
sensors, which allows integration with the camera in the scheme of
\cref{fig:concept}b. 
We characterize the optical response of each
projector by using linearly polarized light with wavelengths from 
\SIrange{400}{1000}{nm}. Figure 3 in the Supplementary Material shows the
experimentally measured responses of the metasurfaces, illustrating excellent
agreement with the expected theoretical responses. We utilize the fabricated
projector as a fixed encoder to optimize the reconstruction ability of the
neural network decoders. 

\subsection{Spectral Reconstruction} We perform
spectral reconstruction from the barcodes obtained from both the theoretical and
experimental responses of the fabricated metasurface projectors.
\Cref{fig:hardware}b shows a scene from the FVgNET dataset as perceived through
each of the projectors based on experimental data. In \cref{fig:hardware}c we
present a qualitative comparison between the hyperspectral reconstruction of
this scene based on both the simulated and experimental barcodes against the
original. \Cref{fig:hardware}d illustrates a quantitative comparison between the
original spectra and its reconstructions as obtained from the experimental
implementation Hyplex\texttrademark{} and the algorithm by Nie \etal.
\cite{nie2018deeply}. The reconstruction is carried out through the use of the
connected MLP decoder introduced in  \cref{sec:method}. We designate 80\% of our
dataset for training the decoder and the remainder for validation purposes.

\Cref{tab:tablemerit} presents a performance comparison of
Hyplex\texttrademark{} against state-of-the-art reconstruction approaches. We
present the results of the reconstruction from the datasets described in
\cref{sec:datasets}, as well as for the validation part of our own dataset. For
the consistency of the comparison, we adapted the metrics and data reported in
\cite{nie2018deeply}, where the calculated RMSE is normalized into the range
[0, 255] to approximately represent the error in pixel intensity. The
reconstruction error of Hyplex\texttrademark{} is the lowest value among CAVE
and both indoor and outdoor images in the Harvard dataset, showing superior
performance against all state-of-the-art models. We further tested our model on
the KAUST dataset and FVgNET dataset by using the optical response of the
fabricated metasurfaces.

\subsection{Hyperspectral Semantic Segmentation}

\begin{figure*}[htbp] \centering
	\includegraphics[width=\textwidth]{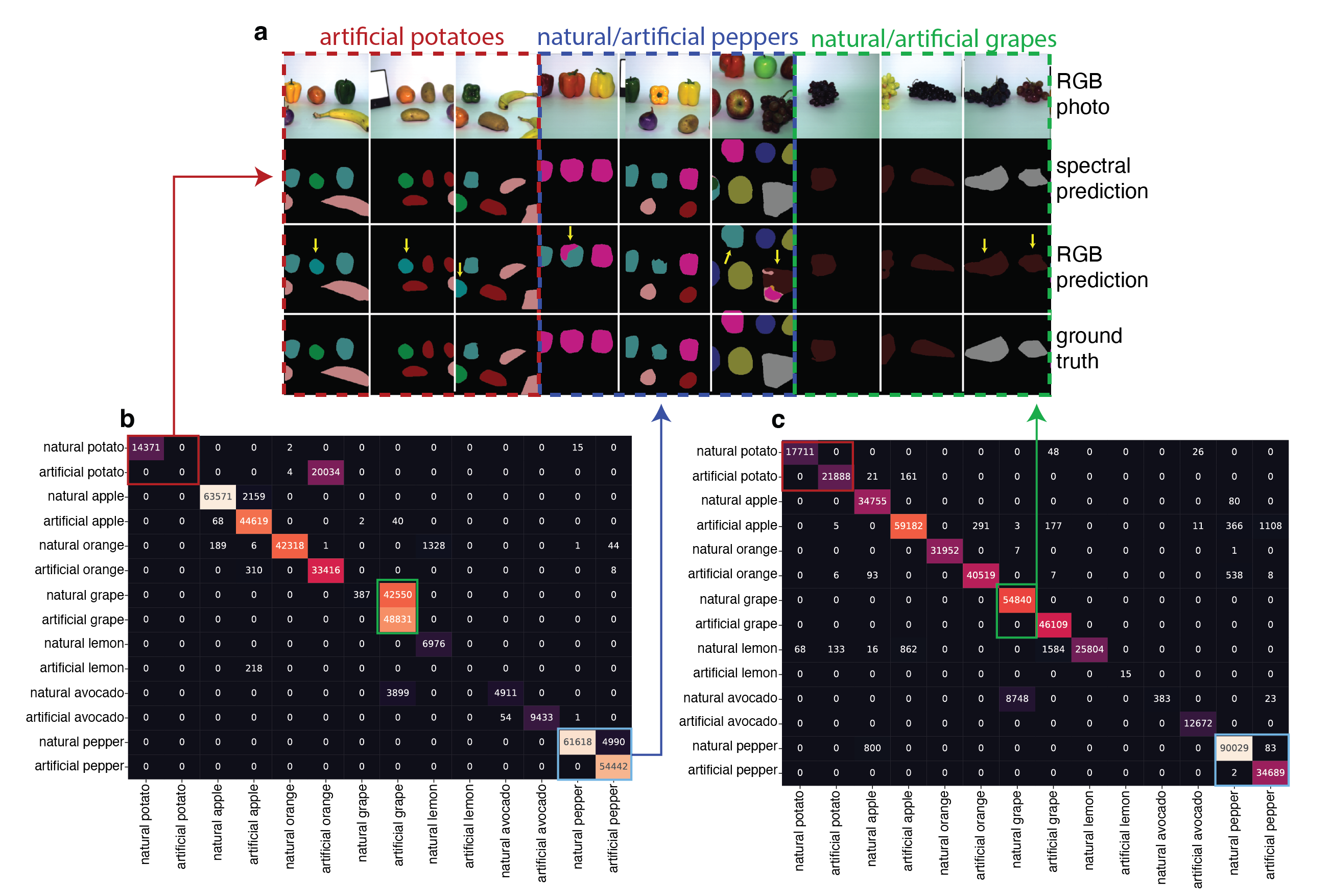}
	\caption{\textbf{Spectral and RGB-based semantic segmentations. }(a) Comparison
		between segmentation masks generated from a spectral-informed model, an RGB-only
		model, and the ground truth. (b) Confusion matrix for RGB only model.  (c)
		Confusion matrix for the spectral-informed model. Each value in the confusion
		matrix represents the number of pixels of the segmentation mask of the item in
		the column that was classified as the item in the row.} \label{fig:segmentation}
\end{figure*}

Here we present labeling of artificial and real fruits from scenes of the FVgNET
dataset. Artificial and real fruits have similar RGB colors. However, they
differ significantly in their reflection spectra. Supplementary Fig.~4 provides
an example of this. We showcase the learning ability of the proposed physical
encoders by training two classification networks. One model uses the spectral
encoders for semantic segmentation labeling, and the second the RGB channels.
Both models use an identical U-Net-like decoder and identical parameters (number
of epochs, batch size, learning rate). The results are summarized in
\cref{fig:segmentation}, where the panel a shows a qualitative comparison of
the segmentation prediction quality for both models against the ground-truth
mask.

\setlength\tabcolsep{14pt} \begin{table}[t] \centering
	\begin{tabular}{@{}lll@{}} \toprule & RMSE & mIoU \\ \midrule Simulation  & 4.23
		& 0.812 \\ Experiment  & 5.41 & 0.741 \\ \bottomrule \end{tabular}
	\caption{\textbf{Simulation and experiment results.} We report RMSE and mIoU
		seperately for reconstruction and segmentation tasks} \label{tab:experiment}
\end{table}

\begin{table}[t] \centering \resizebox{\linewidth}{!}{%
		\begin{tabular}{@{}lllll@{}} \toprule \multicolumn{1}{c}{\multirow{2}{*}{Object
					class}} & \multicolumn{2}{c}{\begin{tabular}[c]{@{}c@{}}Hyperspectral\\
					segmentation\end{tabular}} & \multicolumn{2}{c}{\begin{tabular}[c]{@{}c@{}}RGB\\
					segmentation\end{tabular}} \\ \cmidrule(l){2-5} \multicolumn{1}{c}{} &
			\multicolumn{1}{c}{IoU} & \multicolumn{1}{c}{F1} & \multicolumn{1}{c}{IoU} &
			\multicolumn{1}{c}{F1} \\ \midrule 
			real orange & \textbf{0.979} & \textbf{0.989} & 0.935 & 0.966 \\ artificial
			orange & \textbf{0.954} & \textbf{0.976} & 0.609 & 0.757 \\ real grape &
			\textbf{0.829} & \textbf{0.907} & 0.009 & 0.017 \\ artificial grape &
			\textbf{0.897} & \textbf{0.946} & 0.494 & 0.661 \\ \bottomrule \end{tabular}%
	} \caption{\textbf{Quantitative comparison.} We report 4 examples of object
		classes segmented with HSI and RGB images.} \label{tab:rgbvsspectral}
\end{table} 

While the mask quality is similar for both methods, the mean Intersection over
Union (IoU) score for the spectral-informed model is significantly higher
compared to the RGB one. The mIoU computed with the theoretical and experimental
responses of encoders reaches 81\%, and 74\%, as shown in \cref{tab:experiment}.
With the RGB model, conversely, the mIoU decreases to 68\%.
The confusion matrix of the RGB trained model shows that the RGB model struggles
to predict correct results for real-artificial pairs of fruits with similar
colors (\cref{fig:segmentation}b). The spectral-informed model, conversely,
generates correct labels for most real-artificial pairs
(\cref{fig:segmentation}c) and outperforms the RGB model in IoU and F1
(\cref{tab:rgbvsspectral}). These results demonstrate that the small-sized
barcodes generated by Hyplex\texttrademark{} efficiently compress spectral
features that convey key information about the objects imaged. Table 1 and 2 in
Supplementary Material provide detailed metrics for each object type (apple,
potato, etc.) on both models.

\section{Discussion and Limitations} In this work, we designed and implemented
Hyplex\texttrademark{}, a new hardware system for real-time and high-resolution
hyperspectral imaging. We validated Hyplex\texttrademark{} against current
state-of-the-art approaches and proved it to be outperforming in all benchmarks.
Additionally, we demonstrated the superiority of hyperspectral features and
trainable encoders by designing a model for spectral-informed semantic
segmentation and comparing its performance against RGB models.

One of the limitations in the current implementation of Hyplex\texttrademark{}
is the linear structure of the physical encoder. The study of nonlinear
encoders~\cite{Alameda-Pineda_2016_CVPR} could enable more complex feature
embeddings. This topic may stimulate future research that could generalize the
Hyplex\texttrademark{} framework to include nonlinear metasurfaces, an essential
area of research in the field of
meta-optics~\cite{Maier2018dielectric,Kivshar2018alldielectric}. The second area
of improvement is the spectral sparsity assumption at the core idea of efficient
dimensionality reduction. While this assumption is practically verified in the
majority of computer vision problems~\cite{PV2019102797, Zhang2018}, it may not
hold for specialized tasks. Fabrication errors are also an essential aspect
that, if not adequately considered, can limit performance. In this work, we
mitigate this effect by tuning the software decoder to best use the experimental
response of the projectors. Future work could investigate techniques from
robustness control in inverse design, a new promising area of research
~\cite{Makarenko2021robust,kuhne2021fabrication}. 

Improved results could also be obtained if we augment the publicly
available hyperspectral datasets 
with more scenes obtained at different wavelengths and in different settings
such as, e.g., medical. Such study could generalize the results of
Hyplex\texttrademark{} to provide high impact systems for personalized
healthcare and precision medicine. Hyplex\texttrademark{} could provide a
game-changer technology in this field, leveraging its vast capacity to
fast-process high-resolution hyperspectral images  (see Sec. 7 of Supplementary Material) at speed comparable with current RGB cameras.

\noindent\textbf{Acknowledgements.} This work was supported by the King
Abdullah University of Science and Technology (KAUST) through the Artificial Intelligence Initiative (AII) funding. This research received funding from KAUST (Award OSR-2016-CRG5-2995). Parallel
simulations are performed on KAUST’s Shaheen supercomputer.

\newpage \clearpage {\small \bibliographystyle{ieee_fullname}
	\bibliography{ms} }

\end{document}


\title{Supplementary information for: Real-time hyperspectral imaging in hardware via trained metasurface encoders
}

\author{M. Makarenko$^{\dag}$, A. Burguete-Lopez, Q. Wang, F. Getman, Silvio Giancola, \\Bernard Ghanem \& A. Fratalocchi\\
King Abdullah University of Science and Technology (KAUST)\\
Thuwal, 23955-6900, KSA\\
{\tt\small $^{\dag}$maksim.makarenko@kaust.edu.sa}
}

\maketitle


\section{Details on time-domain coupled-mode theory TDCMT}

\begin{figure}[htbp]
    \centering
    \includegraphics{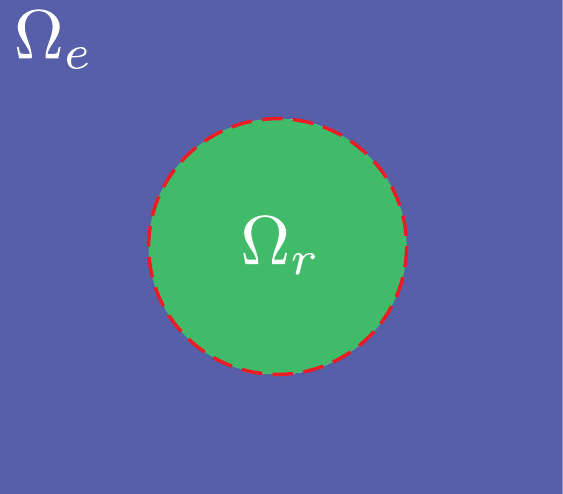}
    \caption{Splitting of metasurface geometry space $\Omega$ into resonance space $\Omega_r$ and exterior (propagation) space $\Omega_e$}
    \label{sup_fig1}
\end{figure}

In this section, we describe a set of exact coupled-mode equations that
are fully equivalent to Maxwell's equations. The main idea of the coupled-mode approach is to divide the geometrical space $\Omega$ where light propagation into a resonator space $\Omega_{r}$ and an external space $\Omega_{e}$ (Fig. \ref{sup_fig1}). We assume that the external space does not contain sources or charges. Under this formulation, the set of Maxwell equations reduces to the following set of exact coupled-mode equations \cite{Makarenko2020}:

\begin{equation}\label{supp_cme}
\left\{\begin{array}{l}
\tilde{\mathbf{a}}(\omega)=\frac{\tilde{K}}{i(\omega-W)+\frac{\tilde{K} \tilde{K}^{\dagger}}{2}} \tilde{\mathbf{s}}_{+} \\
\tilde{\mathbf{s}}_{-}(\omega)=\tilde{C}\left(\tilde{\mathbf{s}}_{+}-\tilde{K}^{\dagger} \cdot \tilde{\mathbf{a}}\right)
\end{array}\right.
\end{equation}
with $1 / \tilde{X}$ the inverse matrix $\tilde{X}^{-1}$. Power conservation implies that the matrix $\sigma$ :

\begin{equation}
\sigma=\mathbf{1}-\tilde{K} \frac{1}{i(\omega-W)+\frac{\tilde{K} \tilde{K}^{\dagger}}{2}} \tilde{K}^{\dagger}
\end{equation}
defined from the solution of the coupled mode equations, is unitary $\sigma^{\dagger} \cdot \sigma=1$.

Equations \eqref{supp_cme} show that the dynamics of the system depend only on three independent matrices: the coupling matrix $\tilde{K}$, the scattering matrix $\tilde{C}$, and the resonance matrix $W$. 

\section{Network training for supervised spectral prediction}

\begin{figure}[htbp]
    \centering
    \includegraphics[width=\linewidth]{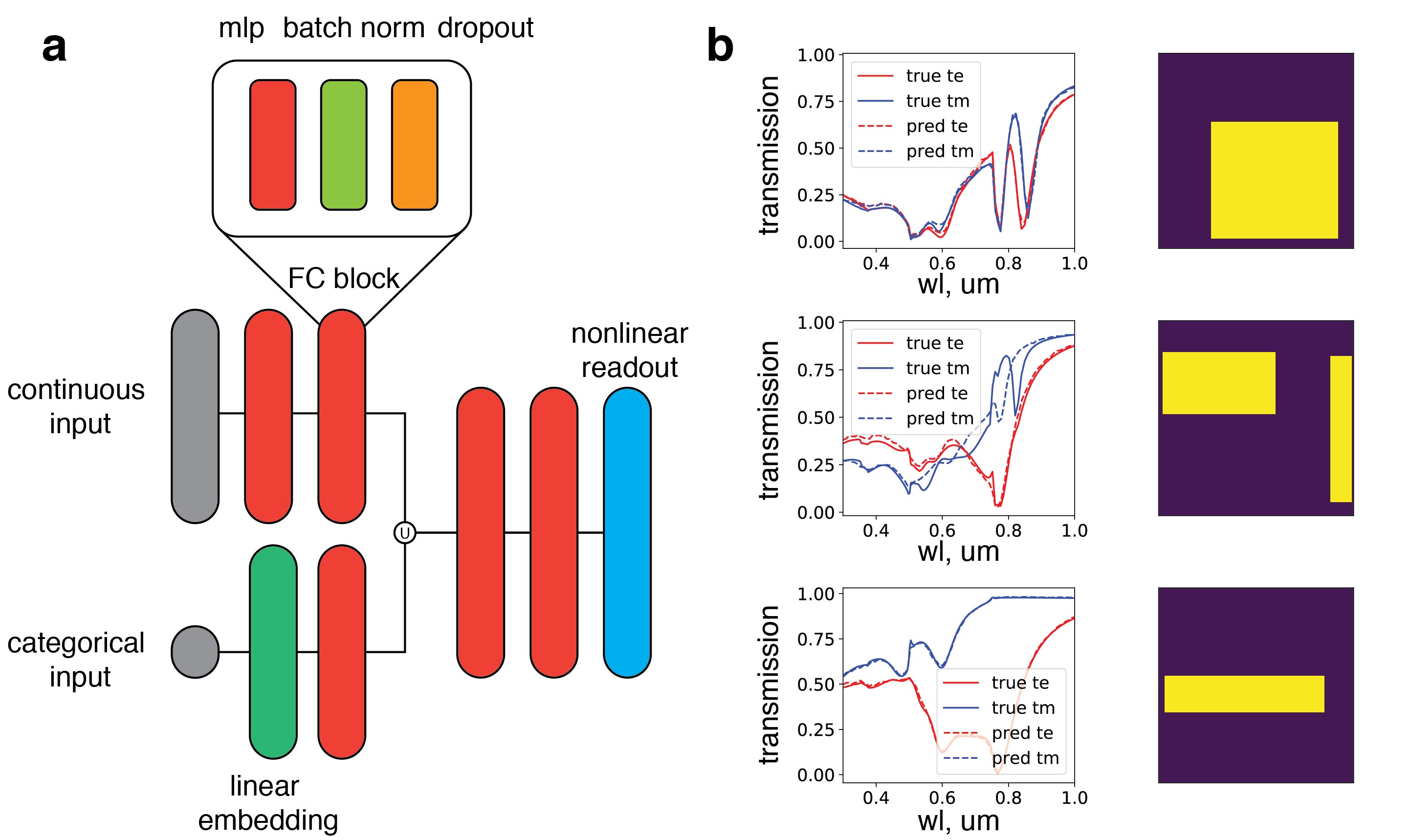}
    \caption{Differentiable ALFRED spectral predictor. \textbf{a} Conceptual sketch of d-alfred neural network shape-to-resonance mapper. \textbf{b} Qualitative results of spectra prediction for the dataset samples.}
    \label{sup_fig2}
\end{figure}

Figure \ref{sup_fig2} illustrates the model of the proposed differentiable spectral predictor (d-ALFRED). It consists of several fully connected (FC) blocks connected sequentially. Each FC block consists of multi-layer perceptrons (MLP) of different sizes, batch normalization layer, and dropout. To process separately categorical variables in input (period, thickness), we design an alternative branch consisting of a linear embedding layer and FC block connected sequentially. The primary purpose of this branch is to balance categorical and continuous variable's weights in the model. Then we concatenate both continuous and categorical features and feed them into readout blocks, consisting of multiple FC blocks. 

We use the training dataset provided by \cite{Makarenko2021robust}, which contains over \num{600000} simulation results of pure silicon structures on top of glass under a Total-Field Scattered-Field (TFSF) simulation. Each simulation has periodic boundary conditions with one of the three different periods (\SI{250}{nm}, \SI{500}{nm} or \SI{750}{nm}) and one of the ten different discrete thicknesses from \num{50} to \SI{300}{nm} with a \SI{25}{nm} step. Each structure consists of a random combination (up to 5) of cuboid resonators. We split the dataset into test and training parts comprising 20\% and 80\% of the total, respectively, then take 10\% of the training set as a validation set.

For the training part, we use the Adam optimizer \cite{adam} with a learning rate \num{1e-5} and a step learning rate scheduler with $stepsize=50$ and $\gamma=0.1$ hyperparameters. For the desired system response in either transmission or reflection, we apply a sigmoid activation function at the top layer of FCN. This function maps the output spectrum to the range [0,1], which is beneficial for convergence at the beginning of the training stage. Since we use periodic boundary conditions, we used random translation and rotations as data-augmentation.

As a result, we obtain $0.008$ validation mean squared error, which is slightly higher than the previous result with convolution-based model \cite{Makarenko2021robust}. Figure \ref{sup_fig2} provides a qualitative comparison between trained and ground truth spectral responses. 

\section{Dataset details}

In the main dataset we provide hyperspectral images of real and artificial fruits. The miscellaneous class corresponds to fruits and vegetables that do not have a natural counterpart. Approximately 40\% of the scenes consist of a single row of objects located at the camera's focal plane. The remaining scenes show two rows of objects, with the focal plane located in between. We keep the position of the white reference panel approximately constant throughout the dataset for easy normalization. The hyperspectral images have a spatial resolution of 512$\times$512 pixels and 204 spectral bands. We also provide an RGB image as seen through the lens of the camera for each scene with the same spatial resolution.

\begin{figure}[htbp]
    \centering
    \includegraphics[width=\columnwidth]{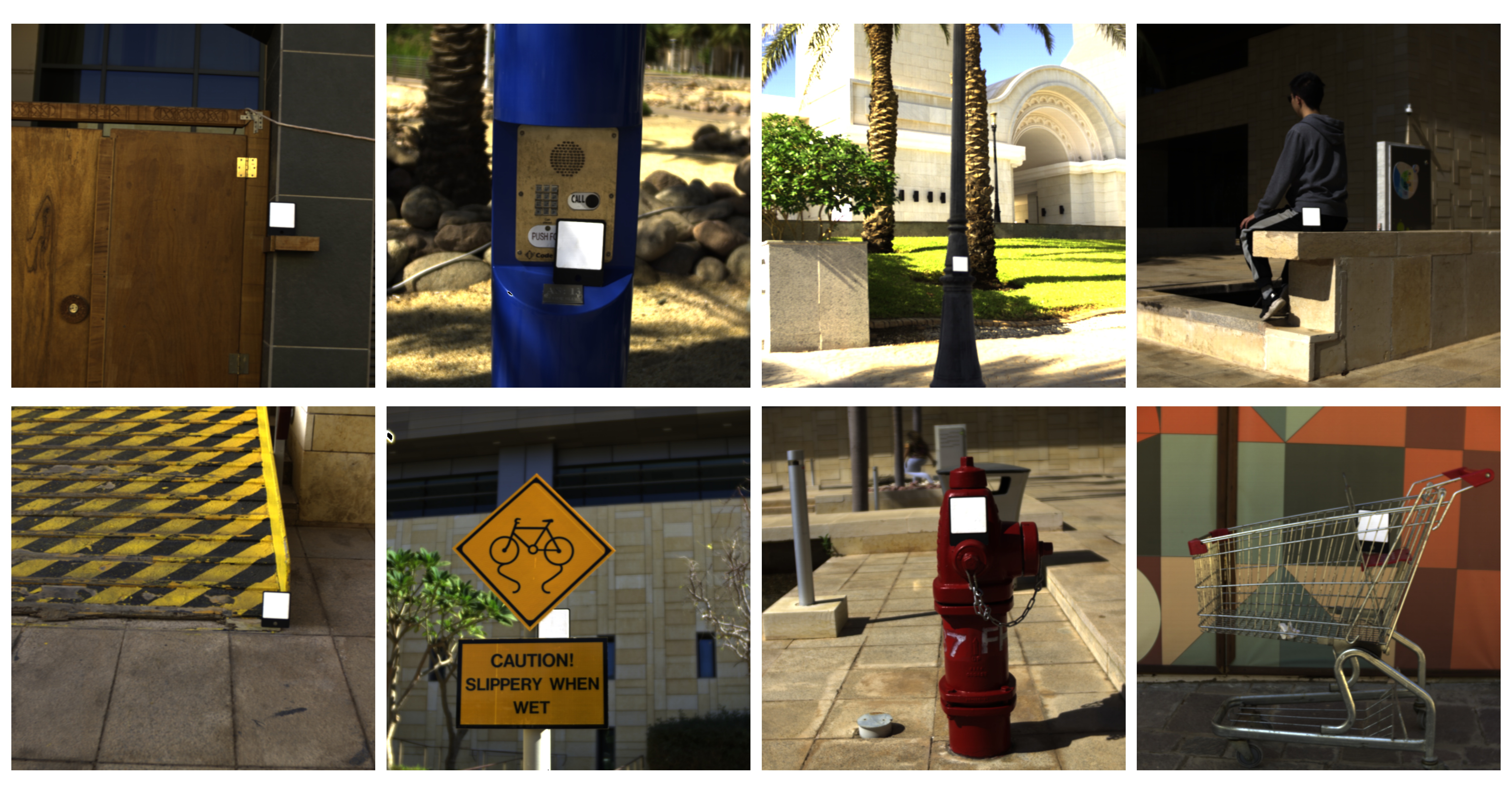}
    \vspace{-.7cm}
    \caption{ \textbf{Additional samples captured in a real-world setting.} }    
    \label{fig:realdataset}
    \end{figure}

To validate the generalization ability of our framework, we augmented the dataset with 20 additional images in the wild (examples can be seen in Fig.~\ref{fig:realdataset}). The resulting reconstruction error for these images is $2.54 \pm 2.72$, a value consistent with the results obtained with the KAUST dataset used to train the encoder. 

\section{Nanofabrication details}
We produce the devices using \SI{15}{mm} wide and \SI{0.5}{mm} thick square pieces of fused silica glass as the substrate. Using plasma-enhanced vapor deposition, we deposit a thin layer of amorphous silicon on the glass, the thickness of which is controlled on each sample to match the design requirements. We then spin coat \SI{200}{nm} of the resist ZEP-520A (ZEON corporation and \SI{40}{nm} of the resist AR-PC 5090 (ALLRESIST) and pattern the shapes of the nanostructures using an electron beam lithography system with a \SI{100}{kV} acceleration voltage. Following this, we remove the AR-PC 5090 by submersing each sample for \SI{60}{s} in deionized water. We develop the samples by submerging them in ZED-50 (ZEON corporation) for \SI{90}{s} and rinse for \SI{60}{s} in isopropyl alcohol. We then deposit \SI{22}{nm} of chromium using electron beam evaporation to create a hard mask and perform liftoff followed by ultrasonic agitation for \SI{1}{\min}. Following this, we remove the unprotected silicon using reactive ion etching, submerge the devices in TechniEtch Cr01 (Microchemicals) for \SI{30}{s} to remove the metal mask, and rinse with deionized water to obtain the final device.

\section{Characterization}
\begin{figure*}
    \centering
    \includegraphics[width=\linewidth]{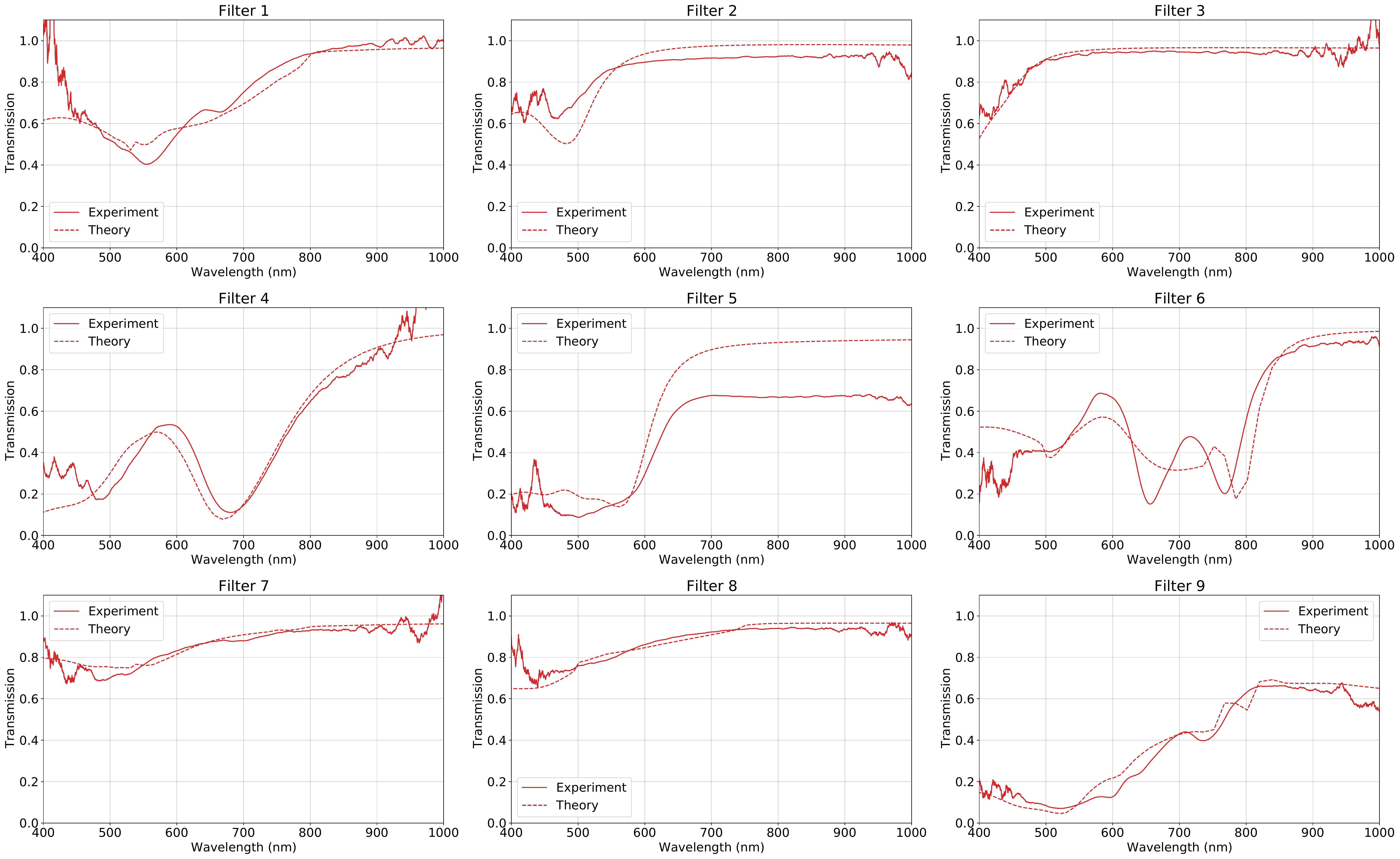}
    \caption{Comparison between theoretical and experimental response of designed filters.}
    \label{fig:filter_responses}
\end{figure*}

We measure the spectral response of our devices in transmission. For accurate characterization, we fabricate each filter separately as a uniform square \SI{500}{\micro\meter} wide. A setup with two 10x microscope objectives allows us to focus broadband (\SI{400}{nm}-\SI{1000}{nm}) linearly polarized light on our samples. A spectrometer then processes the transmitted light. The resulting transmission curves, and their comparison to the theoretical ones, are shown in Fig. \ref{fig:filter_responses}. 
\section{Additional results}

\begin{figure*}[ht]
\centering
    \includegraphics[width=.95\linewidth]{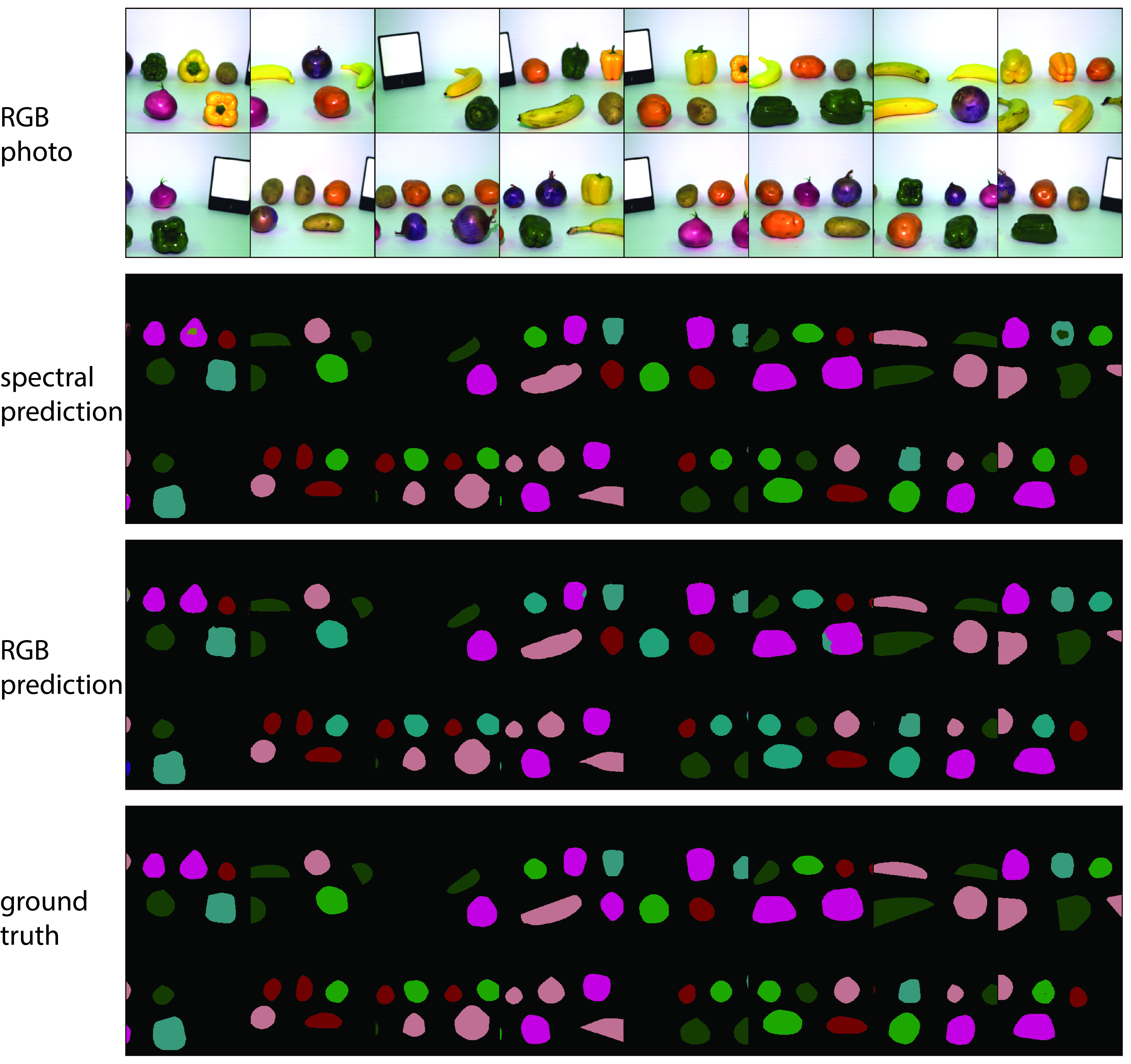}
\caption{Comparison between RGB and spectral-informed models on semantic fruit segmentation task}
    \label{sup:segment}
\end{figure*}

\begin{figure*}[ht]
\centering
    \includegraphics[width=.95\linewidth]{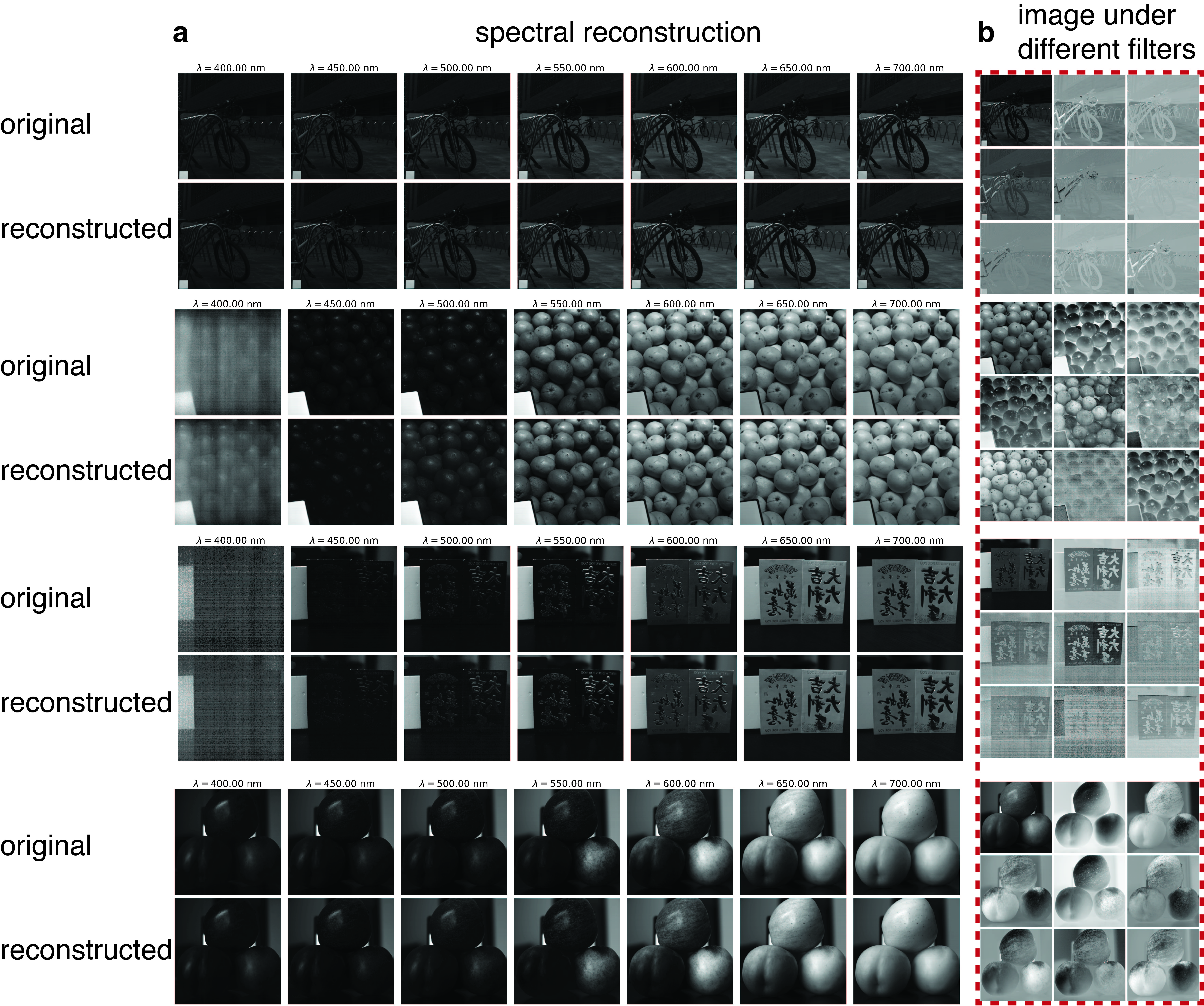}
\caption{Simulation reconstruction results on KAUST dataset. \textbf{a} Image spectral recovery at different wavelengths. \textbf{b} Simulated barcode of the scene as it would be perceived by Hyplex\texttrademark{} through each of the nine projectors.}
    \label{sup:recon}
\end{figure*}

In this section, we provide additional computational results.
Fig. \ref{sup:segment} provides additional qualitative comparisons between RGB trained and Hyplex\texttrademark{} model on the segmentation quality on the FVgNET dataset. 
Fig. \ref{sup:recon} illustrates reconstruction efficiency in simulations of Hyplex\texttrademark{} on the samples from KAUST dataset.

\section{Real-time processing}
The ``first layer'' of the learning model is purely optical and acquires data at the speed of light. 
Therefore, the data acquisition speed of Hyplex\texttrademark{} is limited only by the sensor frame rate (30 FPS in this work). For real-time classification/segmentation tasks, the remaining layers of the network will create delays between the real-time processing of the hyperspectral images and the output for the task. We chose a shallow network implemented in a GPU in this work, which resulted in real-time ($>20$ 
FPS) processing. For better validation purposes, we match the specifications of the dataset we used in 
training and designed the system to work from \SIrange{400}{700}{nm} with \SI{10}{nm} spectral 
resolution and $512\times 512$ spatial resolution. In general, our spectral resolution can achieve up to 
\SI{2}{nm}, covering the wavelength range from \SIrange{400}{700}{nm}. Using a high-resolution camera 
sensor currently available in the market ($>12$ MP), we could produce a $>2$ MP hyperspectral camera 
with an acquisition speed close to \SI{1}{Tb/s}. We will provide these additional details in the suppl. 
material.

.

\begin{table}[htbp]
\centering
\resizebox{\linewidth}{!}{%
\begin{tabular}{@{}llllll@{}}
\toprule
Validation stats   & IoU    & F1     & Prec   & recall & Acc    \\ \midrule
background         & 0.9940 & 0.9970 & 0.9966 & 0.9974 & 0.9948 \\
real potato        & 0.9626 & 0.9809 & 0.9877 & 0.9743 & 0.9999 \\
artificial potato  & 0.9655 & 0.9824 & 0.9877 & 0.9772 & 0.9999 \\
real apple         & 0.9522 & 0.9754 & 0.9645 & 0.9867 & 0.9997 \\
artificial apple   & 0.9021 & 0.9485 & 0.9714 & 0.9267 & 0.9988 \\
real orange        & 0.9786 & 0.9891 & 0.9948 & 0.9836 & 0.9999 \\
artificial orange  & 0.9541 & 0.9764 & 0.9866 & 0.9666 & 0.9996 \\
real grape         & 0.8294 & 0.9067 & 0.8530 & 0.9677 & 0.9978 \\
artificial grape   & 0.8971 & 0.9457 & 0.9331 & 0.9588 & 0.9990 \\
real lemons        & 0.8673 & 0.9289 & 0.9881 & 0.8764 & 0.9992 \\
artificial lemons  & 0.0009 & 0.0018 & 0.7143 & 0.0009 & 0.9968 \\
real avocado       & 0.0365 & 0.0704 & 1.0000 & 0.0365 & 0.9981 \\
artificial avocado & 0.9436 & 0.9709 & 0.9764 & 0.9656 & 0.9999 \\
real pepper        & 0.9586 & 0.9788 & 0.9808 & 0.9769 & 0.9993 \\
artificial pepper  & 0.9426 & 0.9704 & 0.9583 & 0.9828 & 0.9996 \\
real unknown       & 0.9243 & 0.9606 & 0.9491 & 0.9726 & 0.9980 \\
artificial unknown & 0.7016 & 0.8246 & 0.7215 & 0.9621 & 0.9957 \\\midrule
total              & 0.8124 & 0.8476 & 0.9391 & 0.8537 & 0.9986 \\
total(-background) & 0.8011 & 0.8382 & 0.9355 & 0.8447 & 0.9988 \\ \bottomrule
\end{tabular}
}
\caption{Validation stats on spectral segmentation}
\label{Validation_stats1}
\end{table}

\begin{table}[htbp]
\centering
\resizebox{\linewidth}{!}{%
\begin{tabular}{@{}llllll@{}}
\toprule
Validation stats & IoU & F1 & Prec & recall & Acc \\ \midrule
background & 0.9950 & 0.9975 & 0.9968 & 0.9983 & 0.9957 \\
real potato & 0.9688 & 0.9841 & 0.9980 & 0.9707 & 0.9999 \\
artificial potato & 0.0000 & 0.0000 & 0.0000 & 0.0000 & 0.9961 \\
real apple & 0.9469 & 0.9727 & 0.9860 & 0.9598 & 0.9993 \\
artificial apple & 0.9186 & 0.9575 & 0.9301 & 0.9867 & 0.9992 \\
real orange & 0.9351 & 0.9664 & 0.9940 & 0.9404 & 0.9994 \\
artificial orange & 0.6094 & 0.7572 & 0.6239 & 0.9633 & 0.9959 \\
real grape & 0.0088 & 0.0174 & 0.9949 & 0.0088 & 0.9917 \\
artificial grape & 0.4935 & 0.6608 & 0.5015 & 0.9687 & 0.9904 \\
real lemons & 0.8241 & 0.9035 & 0.8387 & 0.9794 & 0.9997 \\
artificial lemons & 0.0000 & 0.0000 & 0.0000 & 0.0000 & 0.9973 \\
real avocado & 0.5467 & 0.7069 & 0.9891 & 0.5500 & 0.9992 \\
artificial avocado & 0.9708 & 0.9851 & 0.9992 & 0.9716 & 0.9999 \\
real pepper & 0.9068 & 0.9511 & 0.9954 & 0.9107 & 0.9988 \\
artificial pepper & 0.8945 & 0.9443 & 0.9081 & 0.9836 & 0.9988 \\
real unknown & 0.9524 & 0.9756 & 0.9762 & 0.9751 & 0.9989 \\
artificial unknown & 0.7274 & 0.8421 & 0.7605 & 0.9435 & 0.9967 \\ \midrule
total & 0.6882 & 0.7425 & 0.7937 & 0.7712 & 0.9975 \\
total(-background) & 0.6690 & 0.7265 & 0.7810 & 0.7570 & 0.9976 \\ \bottomrule
\end{tabular}
}
\caption{Validation stats on RGB segmentation}
\label{tab:validation2}
\end{table}


\begin{figure*}[ht]
\centering
    \includegraphics[width=.5\linewidth]{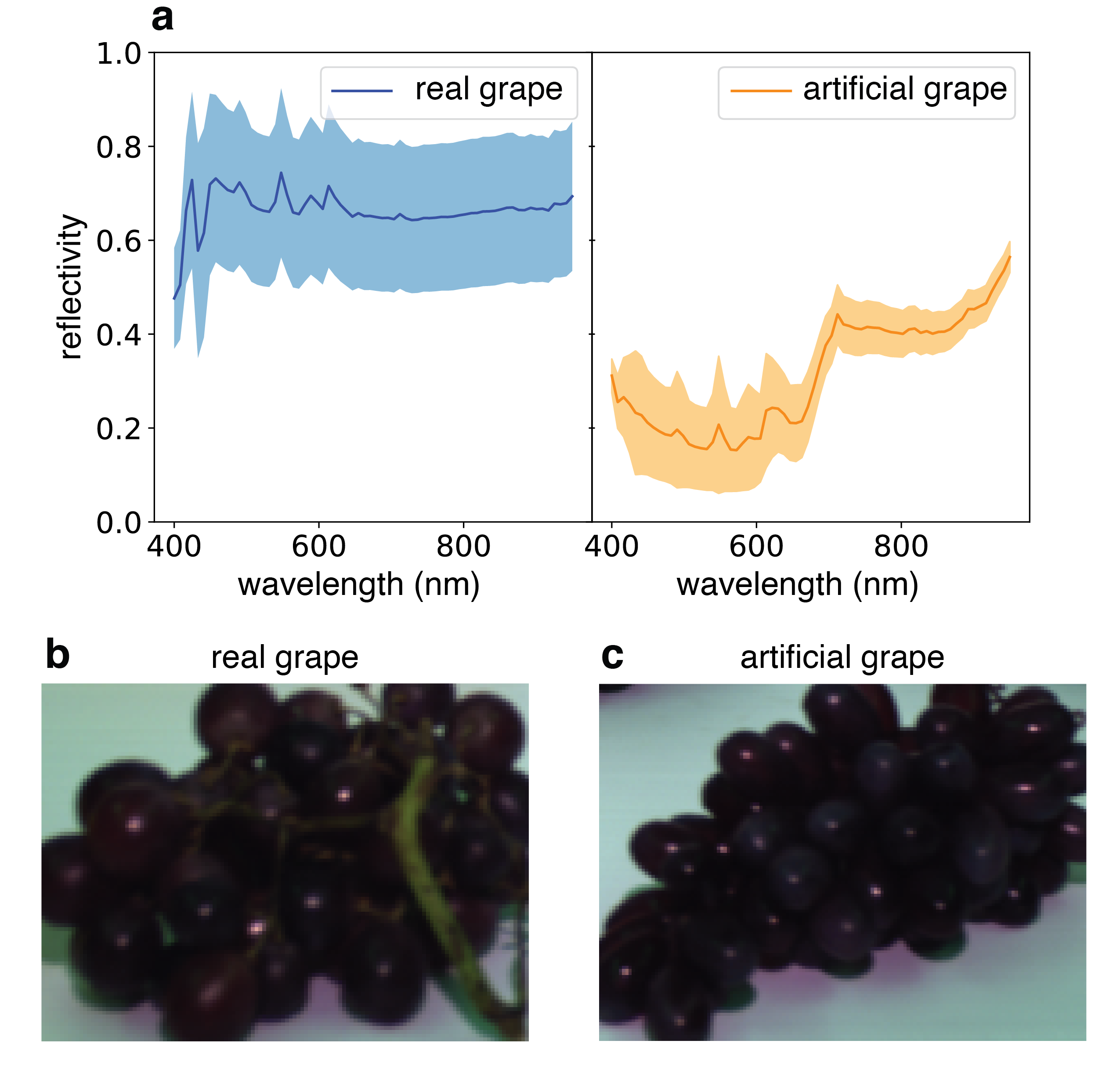}
\caption{Comparison between spectral responses and RGB images of real and artificial grapes.\textbf{a} Reflection spectra of real and artificial grapes. \textbf{b} RGB image of real grapes. \textbf{c} RGB image of artificial grapes.}
    \label{realartificial}
\end{figure*}

\bibliographystyle{ieee_fullname}
\bibliography{cvpr}